\documentclass[10pt,twocolumn,letterpaper]{article}

\usepackage{iccv}              % To produce the CAMERA-READY version

\usepackage{multirow}
\usepackage{color}
\usepackage{algorithm}
\usepackage{algorithmic}
\definecolor{greenn}{rgb}{0.21,0.9,0.21} 
\definecolor{redd}{rgb}{0.9,0.21,0.21}

\definecolor{iccvblue}{rgb}{0.21,0.49,0.74}
\usepackage[pagebackref,breaklinks,colorlinks,allcolors=iccvblue]{hyperref}

\def\paperID{450} % *** Enter the Paper ID here
\def\confName{ICCV}
\def\confYear{2025}

\title{Next Patch Prediction for AutoRegressive Visual Generation}

\author{\textsuperscript{1,3}Yatian Pang\textsuperscript{$\dag$}, \textsuperscript{1}Peng Jin, \textsuperscript{1}Shuo Yang, \textsuperscript{1,7}Bin Lin,  \textsuperscript{1,7}Bin Zhu,  \textsuperscript{1}Zhenyu Tang,  \textsuperscript{1,7}Liuhan Chen,  \\ \textsuperscript{3}Francis E. H. Tay, \textsuperscript{5,6}Ser-Nam Lim, \textsuperscript{4,6}Harry Yang, \textsuperscript{1,2}Li Yuan\thanks{ Corresponding author, yuanli-ece@pku.edu.cn \\  $\dag$ yatian\_pang@u.nus.edu }  
\\
\\
  \textsuperscript{1}Peking University 
  \textsuperscript{2}PengCheng Laboratory 
  \textsuperscript{3}NUS 
  \textsuperscript{4}HKUST 
  \textsuperscript{5}UCF 
  \textsuperscript{6}Everlyn
  \textsuperscript{7}Rabbitpre AI
}

\begin{document}
\maketitle

% \twocolumn[{
% \renewcommand\twocolumn[1][]{#1}
% \maketitle
% \centering
% \includegraphics[width=0.7\linewidth]{teaser.png}

% \captionof{figure}{\textbf{Comparison of our method and baseline methods.} Our method achieves higher FID scores with less training cost on the ImageNet benchmark.}
% \label{fig:teaser}

% }]

\begin{abstract}
Autoregressive models, built based on the Next Token Prediction (NTP) paradigm, show great potential in developing a unified framework that integrates both language and vision tasks. Pioneering works introduce NTP to autoregressive visual generation tasks. In this work, we rethink the NTP for autoregressive image generation and extend it to a novel \textbf{Next Patch Prediction} (NPP) paradigm. Our key idea is to group and aggregate image tokens into patch tokens with higher information density. By using patch tokens as a more compact input sequence, the autoregressive model is trained to predict the next patch, significantly reducing computational costs. To further exploit the natural hierarchical structure of image data, we propose a multi-scale coarse-to-fine patch grouping strategy. With this strategy, the training process begins with a large patch size and ends with vanilla NTP where the patch size is 1$\times$1, thus maintaining the original inference process without modifications. Extensive experiments across a diverse range of model sizes demonstrate that NPP could reduce the training cost to $\sim 0.6\times$ while improving image generation quality by up to 1.0 FID score on the ImageNet 256$\times$256 generation benchmark. Notably, our method retains the original autoregressive model architecture without introducing additional trainable parameters or specifically designing a custom image tokenizer, offering a flexible and plug-and-play solution for enhancing autoregressive visual generation. \url{https://github.com/PKU-YuanGroup/Next-Patch-Prediction}
\end{abstract}

\section{Introduction}
\begin{figure}[h]
    \centering
    \includegraphics[width=\linewidth]{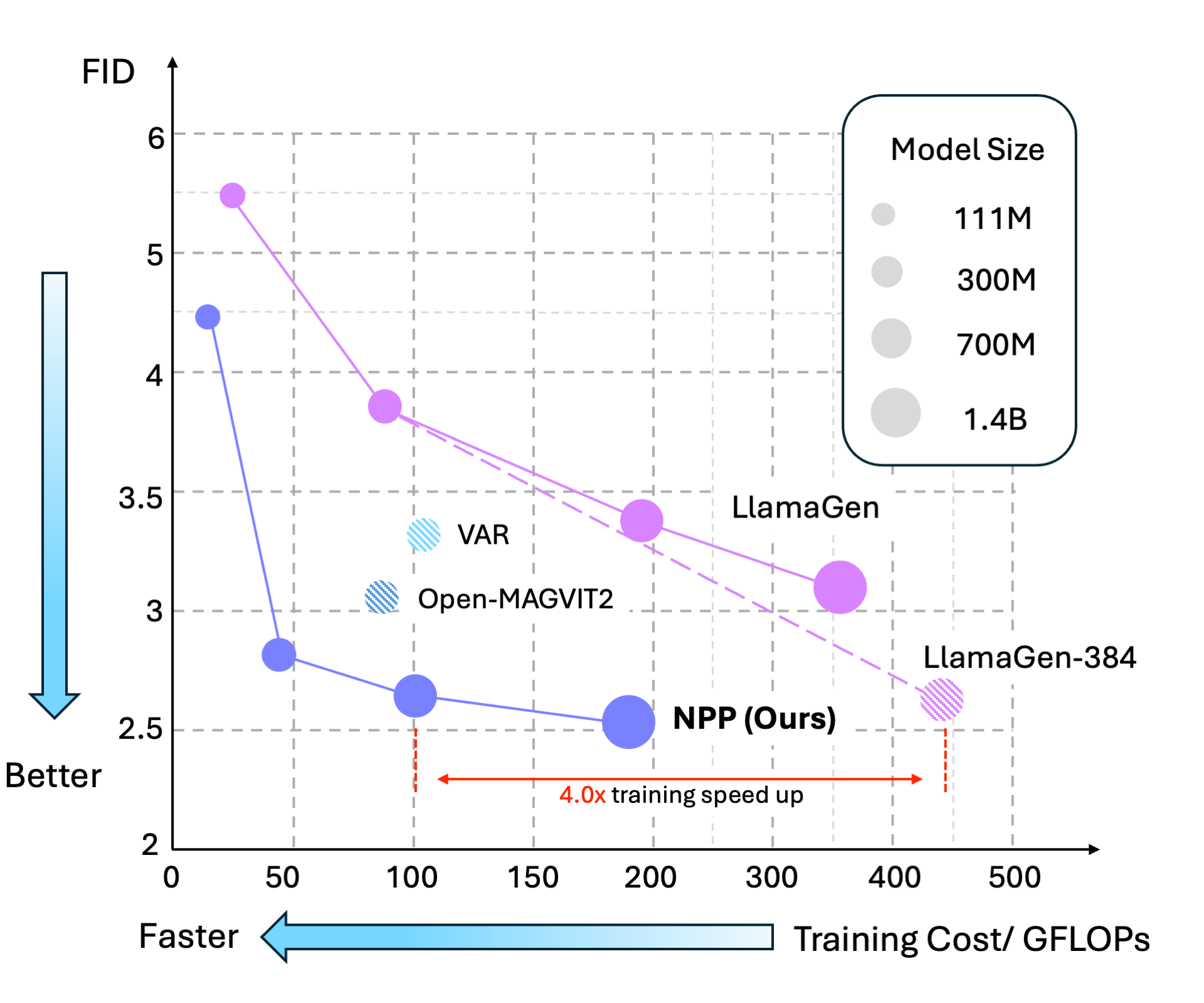}
    % \vspace{-1em}
    \caption{\textbf{Comparison of our method and baseline methods.} Our method on a diverse range of models achieves higher FID scores with significantly less training cost on the ImageNet 256$\times$256 generation benchmark. Our method NPP-L achieves up to $4.0 \times$ training speed up without performance degradation compared to LlamaGen-L-384.}
    \label{fig:teaser}
\end{figure}
Autoregressive models, foundational to large language models (LLMs)~\citep{transformer,bert, gpt1,t5,gpt2,gpt3,opt}, generate content through the prediction of subsequent tokens in a sequence. This Next Token Prediction (NTP) paradigm enables LLMs to excel in a variety of natural language processing tasks, exhibiting human-like conversational abilities~\citep{gpt3.5,openai2022chatgpt,gpt4,google2023bard,anthropic2023claude,bloom,Llama1,Llama2,qwen,baichuan,internlm,deepseek} and demonstrating remarkable scalability~\citep{scalinglaw,scalingar,chinchilla,emergent,revisitingscalinglaws,palm,palm2}. Such advancements illustrate the potential for achieving general-purpose artificial intelligence systems. 
\begin{figure*}[h]
    \centering
    \includegraphics[width=\linewidth]{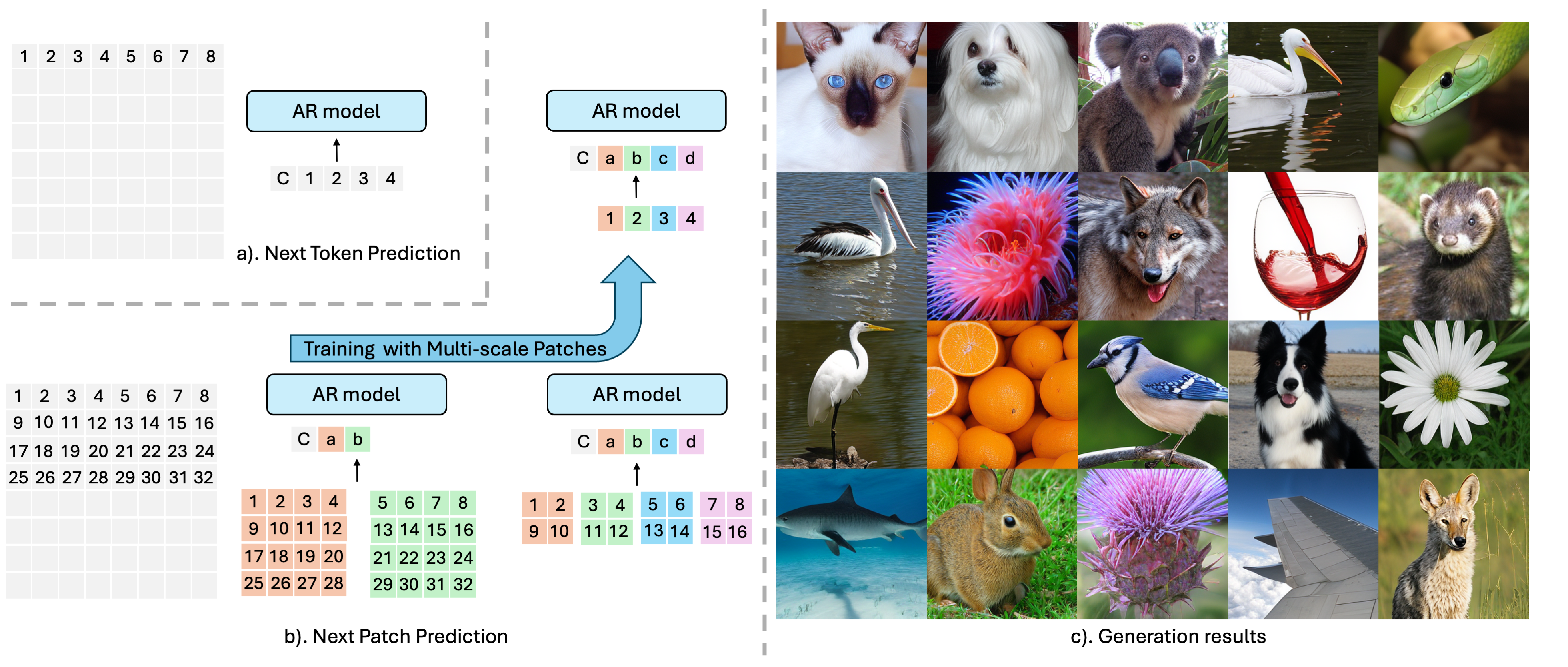}
    % \vspace{-1em}
    \caption{\textbf{Motivation of the next patch prediction.} a). Illustration of next token prediction. b). Demonstration of the proposed next patch prediction. c). Generation results on the ImageNet benchmark. Please zoom in to view.}
    \label{fig:motivation}
\end{figure*}
Inspired by the success of autoregressive models in the language domain, their applications for image generation have been widely %are significantly 
explored. Notable approaches, including VQVAE~\citep{vqvae, vqvae2}, VQGAN~\citep{vqgan, rq}, DALL-E~\citep{dalle1}, and Parti~\citep{vit-vqgan, parti}, introduce image tokenizers that convert continuous images into discrete tokens, employing autoregressive models to sequentially generate these tokens, thereby achieving image generation. In parallel, diffusion models~\citep{scorebased,ddpm,ddim} emerge as a distinct and rapidly evolving paradigm in image generation. However, the fundamental differences in the underlying methodologies of autoregressive and diffusion models pose significant challenges for developing a unified framework that integrates both language and vision tasks. 

More recently, a pioneering work LlamaGen~\citep{llamagen} achieves the next token prediction paradigm for image generation with a vanilla autoregressive model, Llama, bringing the field one step closer to building a unified model between language and vision. However, directly applying NTP from the language domain to the image domain may lead to suboptimal performance due to the distinct properties of the two different modalities. 

In this work, we follow the NTP paradigm as shown in Figure~\ref{fig:motivation} a) for autoregressive image generation and rethink the modeling of the NTP paradigm in the following aspects. 
\begin{itemize} 
\item The NTP paradigm, widely successful in large language models, leverages the high information density of text tokens. However, image tokens typically exhibit lower information density due to the inherently redundant nature of image data. Our key insight is to aggregate multiple image tokens into high information density units referred to as patches\footnote{Here, we define the patch contains multiple image tokens originally encoded by the VQVAE encoder.}, which can potentially enhance the performance of autoregressive image generation.
% \snl{Possible question, image tokens are potentially already image patches? So when you say patches in NPP, you mean the token is representing multiple patches?}

\item Transformer-based autoregressive models incur substantial computational costs during training, with the total cost approximately scaling as $C \approx 6WN$~\cite{kaplan2020scaling}, where 
$W$ represents the number of model parameters and 
$N$ denotes the input sequence length. While maintaining the model architecture, we could manage to reduce the input sequence length of image tokens, thus improving training efficiency.

\item Unlike language data, image modality inherently exhibits hierarchical property in both understanding and generation tasks. This observation suggests that autoregressive image generation could benefit from a multi-scale, coarse-to-fine modeling strategy, which has the potential to improve generation quality and training efficiency.

\end{itemize}

Building on these insights, we introduce Next Patch Prediction (NPP)  as shown in Figure~\ref{fig:motivation} b), a simple yet effective method for autoregressive visual generation. Specifically, the input image tokens are grouped and aggregated into patch tokens with higher information density through an intra-patch average operation. With the resulting patch tokens as a shorter input sequence, the autoregressive model is trained to predict the next patch, thus significantly reducing the computational cost.  To further exploit the hierarchical nature of images, we propose a multi-scale patch grouping strategy that progressively refines predictions in a coarse-to-fine manner, seamlessly extending the vanilla NTP paradigm to our novel NPP paradigm. Specifically, the training process starts with a large patch size and ends with vanilla NTP where the patch size is 1$\times$1, thus preserving the original inference stage without requiring modifications. Extensive experiments show that our method not only enhances training efficiency but also improves the generation quality. As shown in Figure~\ref{fig:teaser}, experiments on a diverse range of models from 100M to 1.4B parameters demonstrate that the NPP paradigm could reduce the training cost to $\sim 0.6\times$  while improving image generation quality by up to 1.0 FID score on the ImageNet 256$\times$256 generation benchmark. Some of the generation results are shown in Figure~\ref{fig:motivation} c).
We highlight that our method retains the original autoregressive model architecture without introducing additional trainable parameters or specifically designing a custom image tokenizer. This ensures flexibility for seamless adaptation to various autoregressive models addressing visual generation tasks.

To sum up, this work contributes in the following ways: 

\begin{itemize} 
\item We propose a simple yet effective method to aggregate image tokens into high information density patch tokens. Meanwhile, with patch tokens as a shorter input sequence, our approach enables the autoregressive model to efficiently process and predict the next patch tokens, significantly lowering computational costs.

\item Leveraging the hierarchical property of image modality, we further introduce a multi-scale patch strategy to seamlessly extend the next token prediction paradigm to our novel next patch prediction paradigm.

\item Experiments on a diverse range of models demonstrate that our method could reduce the training cost to $\sim 0.6\times$ while improving image generation quality by up to 1.0 FID score on the ImageNet generation benchmark. 

\end{itemize}
\section{Related Works}

\subsection{Visual Generation}
Generative adversarial networks (GANs)~\citep{gan,biggan,stylegan,gigagan} are the pioneering method for visual generation in the deep learning era, focusing on learning to generate realistic images through adversarial training. Inspired by language model architectures, BERT-style models~\citep{maskgit,muse,magvit,magvit2,weber2024maskbit} emerge, using masked-prediction techniques to learn to predict missing parts of images, much like how BERT predicts masked words in text. Diffusion models~\citep{ddpm,scorebased,ddim,adm,lin2024open,dpm-solver,cascade-diffusion,cfg,ldm,dalle2,imagen,ldm,dit,sdxl,raphael,pixart,gentron,dalle3,playgroundv2.5,sd3} introduce a novel approach, treating visual generation as a reverse diffusion process, where images are gradually denoised from Gaussian noise through a series of steps. Autoregressive models~\citep{vqgan,dalle1,parti}, inspired by GPT, predict the next token in a sequence.  These methods often involve an image tokenization step~\citep{vae, vqvae}, converting pixel space into a more semantically meaningful representation and training the autoregressive model with encoded tokens. Some works~\citep{yu2024an,  zhu2024stabilize, han2024infinity,shi2024taming,luo2024open,chen2024softvqvae,li2024imagefolder} focus on image tokenizer for better compression and reconstruction of image data, which is also crucial for the image generation quality. 

Recently, a pioneering work LlamaGen~\citep{llamagen} introduced the next token prediction paradigm for image generation with a vanilla autoregressive model. VAR~\cite{VAR} proposes a novel next scale prediction, however requiring a specialized multi-scale tokenizer and incurring longer input token sequences. In this work, we follow LlamaGen for autoregressive visual generation and extend the next token prediction paradigm to our novel next patch prediction. Concurrently, a series of works~\citep{ pang2024randar, yu2024randomized, he2024zipar,wang2024parallelized} explore different novel modeling strategies for autoregressive visual generations, including next random token prediction, and parallelized tokens prediction.  However, these works do not focus on training efficiency and largely modify the autoregressive property, inevitably introducing additional complexity to the model. In contrast,  our method focuses on training efficiency and preserves the original autoregressive model architecture without introducing additional trainable parameters or specifically designing a custom image tokenizer.

\subsection{Multimodal Foundation Models}

Recent advancements in large language models and vision-and-language models~\citep{llava, minigpt4, instructblip, kosmos-2, gpt4roi, groma,lin2023video,lin2024moe,jin2024chat,shao2024patch,BLT,the2024large,deepseekai2025deepseekr1incentivizingreasoningcapability,wang2024qwen2, yang2024qwen2} have demonstrated impressive capabilities in various language and vision tasks. However, unifying the understanding and generation tasks in multimodal large language models is still being explored. Most existing approaches~\citep{emu_baai, emu2_baai, dreamllm, seed,zhou2024transfusion,xiao2024omnigen,xie2024showo,li2024autoregressive, gu2024dart,tang2024hart,ma2024janusflow,tong2024metamorph,deng2024nova,hui2025arflow,ren2024flowar} focus on integrating diffusion models with other existing pre-trained models, rather than adopting a unified next-token prediction paradigm. These methods often require complex designs to link two distinct training paradigms, which makes scaling up more challenging and inevitably disconnects visual
token sampling from the multimodal large language models. Some pioneering efforts~\citep{unified-io, unified-io2,lvm,gemini,chameleon,wu2024janus,wang2024emu3,wu2024liquid,zhao2025qlip,qu2024tokenflow,chen2025janus} explore incorporating image generation into large language models using an autoregressive approach, achieving promising results. However, most of them directly adopt the next token prediction paradigm without exploring novel autoregressive visual generation approaches. In this work, our method does not introduce additional trainable parameters or specifically design a custom image tokenizer, ensuring flexibility for seamless adaptation to various autoregressive image generation tasks, including unified vision-language models for understanding and generation tasks.

\begin{figure*}[ht]
    \centering
    \includegraphics[width=0.9\linewidth]{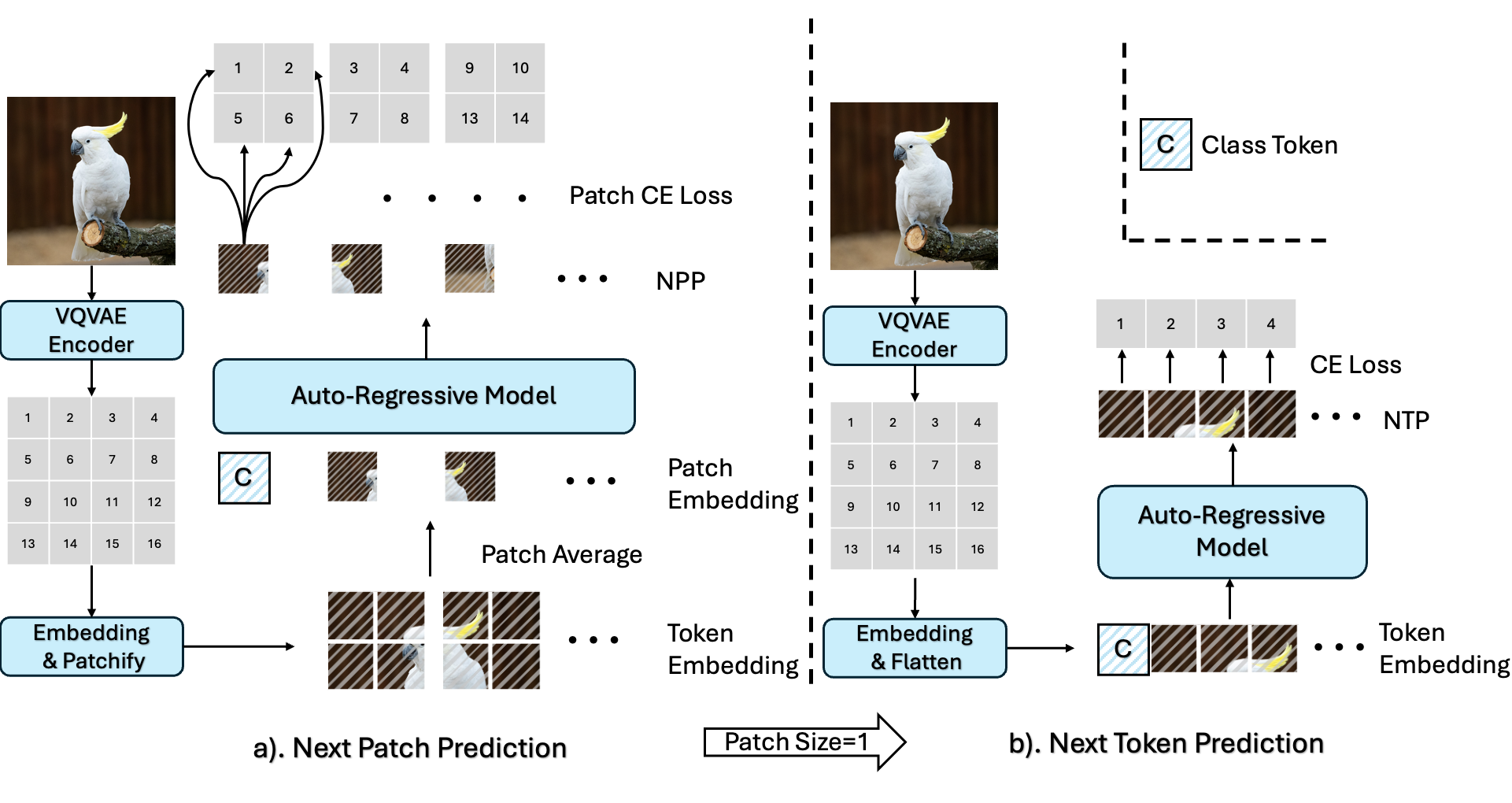}
    \vspace{-1em}
    \caption{\textbf{Next Patch Prediction.} The input image token embeddings are grouped and aggregated into patch embeddings through a path average operation. The autoregressive model is trained to predict the next patch by employing the patch Cross Entropy loss.}
    \label{fig:framework}

\end{figure*}

\section{Method}
In this section, we first provide an overview of the next token prediction paradigm for autoregressive visual generation in Section~\ref{sec:3.1}, followed by our NPP in Section~\ref{sec:3.2}.

\subsection{Preliminaries}
\label{sec:3.1}
We outline the vanilla NTP as shown in Figure~\ref{fig:framework} b). An input image is first encoded into a sequence of discrete tokens $\mathbf{x} = [x_1, x_2, ...,x_K]$ by a pre-trained VQVAE encoder. The autoregressive model is trained to model the probability distribution of a sequence based on a forward autoregressive factorization. Specifically, the training objective is to maximize the joint probability of predicting the current token $x_k$ given the condition token $c$ and all preceding tokens $[x_1, x_2, ...,x_{k-1}]$:

\begin{equation}
% \small
% \vspace{-1em}
\underset{\theta}{\mathrm{max}}\ p_{\theta}(\mathbf{x}) = \prod_{k=1}^Kp_{\theta}(x_k |c, x_1, x_2, \cdots, x_{k-1}),
\label{eq:ar}
% \vspace{-1em}
\end{equation}
where $p_{\theta}$ represents the token distribution predictor with an autoregressive model parameterized by $\theta$. The model utilizes a stack of transformer layers with causal attention, commonly known as a decoder-only transformer. During the inference stage, the model takes a class token as the condition and generates the following image tokens in an autoregressive manner. In this work, we focus on exploring the modeling method for input token sequence and retain
the original autoregressive model architecture without introducing additional trainable parameters or specifically signing a custom image tokenizer.

\subsection{Next Patch Prediction}
\label{sec:3.2}

We introduce the Next Patch Prediction paradigm in Figure~\ref{fig:framework} a). The input image is initially encoded into image token indexes, which are then mapped to token embeddings of sequence length $N$.   Considering the naturally low information density of image data, our key idea is to aggregate multiple tokens into groups of units containing higher information density.  Specifically, we group tokens into non-overlapping patches and generate a sequence of patch embeddings with length $\frac{N}{K}$, where $K$ is the number of tokens associated with each patch.  To avoid introducing extra parameters during this compression process, we simply adopt an intra-patch average operation to compute the patch embeddings. Formally, given the embedding function $E$, for the $i$-th patch $p_i$ associated with $K$ image tokens $x_k^i$ in the input sequence, the patch embedding is formulated as,
\begin{equation}
    E(p_i) = \frac{1}{K}\sum_{k=1}^K E(x_k^i).
\end{equation}

\noindent In this way, the original input token embeddings of sequence length $N$ are aggregated into patch embeddings of sequence length $\frac{N}{K}$. With the resulting patch embeddings as input, the autoregressive model is trained to predict the next patch. However, directly maximizing the joint probability as in Equation~\ref{eq:ar} is difficult due to the absence of an explicit ground truth (GT) index for a patch token. To address this issue, we maintain the original prediction head and propose a patch-wise Cross-Entropy (CE) loss that supervises the model using the associated $K$ image token GT indexes $Index_k^i$ in the next patch $p_i$. Specifically, given the next patch predictions as $Pred_i=P(p_i|c,p_{<i})$, and recalling the patch sequence length $\frac{N}{K}$, the loss function is formulated as:
\begin{equation}
    L = -\frac{1}{N}\sum_{i=1}^\frac{N}{K}\sum_{k=1}^K \log (Pred_i).
\end{equation}

% \snl{i may not fully understand this. The model predicts $Pred_i$, how do you recover the original $x^i_k$ from $Pred_i?$}
%
% Cross-Entropy (CE) Loss cannot be directly applied.maintain the model architecture without introducing additional parameters, 
\begin{figure}[h]
    \centering
    \includegraphics[width=\linewidth]{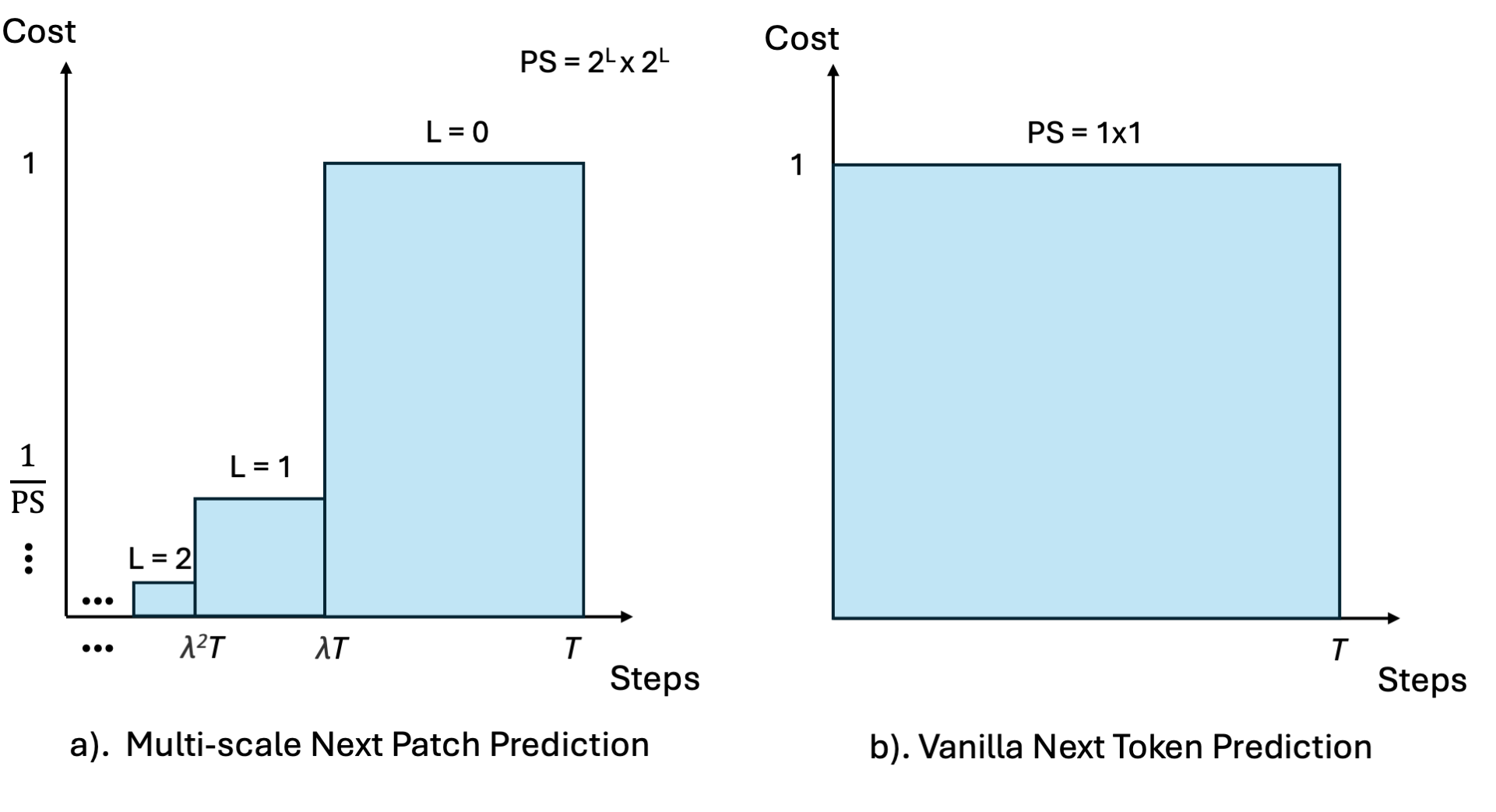}
    \caption{\textbf{Multi-scale Next Patch Prediction.}  The patch grouping function begins with a large patch size, resulting in a short sequence length. As training progresses, the patch size is gradually reduced to $1\times1$.}
    \label{fig:ms}
\end{figure}

However, simply training with this objective leads to the issue that all tokens in a patch are predicted to be the same during inference stage. To address this issue and seamlessly extend the next token prediction
paradigm to our novel next patch prediction paradigm, we propose a multi-scale, coarse-to-fine patch grouping strategy that leverages the natural hierarchical structure of image data as illustrated in Figure~\ref{fig:ms}.  Specifically, the grouping function begins with a large kernel size, resulting in large patches and a short patch sequence length, allowing the autoregressive model to capture coarse representations.  As training progresses, the patch size is gradually reduced to $1\times1$, enabling the model to learn finer details. This strategy seamlessly extends NTP to NPP, making the NPP inference process identical to the vanilla NTP inference stage. To balance training efficiency and model performance, we introduce a segment scheduling factor $\lambda$ and set the number of patch levels $\#L$. During the total training steps $T$, each segment is represented as $\lambda^L T - \lambda^{(L-1)} T$ with a patch size (PS) of $2^L \times 2^L$, where $L$ denotes the current patch level. The computational cost is reduced by a factor of $\frac{1}{PS}$ due to the shorter sequence length at each level.

% \begin{figure}[h]
%     \centering
%     \includegraphics[width=0.6\linewidth]{rope.png}
%     \caption{\textbf{RoPE average.} PS stands for patch size. }
%     \label{fig:rope}
% \end{figure}

To ensure the learned knowledge is transferred smoothly at different patch scales during the training process, we study the effect of Rotary Position Embedding (RoPE)~\cite{su2024roformer} adopted by the autoregressive model. The 2D RoPE embedding $PE$ at image token position $[h_i,w_i]$ can be represented as $PE = RoPE(h_i,w_i)$. Intuitively, when aggregating image tokens into patch tokens, we should also group positions into patches and average them to represent the patch position. However, our pilot study found this design to be unnecessary, so we retain the original form of RoPE for patch token position embeddings $PE = RoPE(h_p,w_p)$, where $[h_p,w_p]$ is the relative patch position.

We present the pseudo code of NPP in Algorithm~\ref{algo:pseduo_code}.

\definecolor{commentcolor}{rgb}{0.2, 0.6, 0.2}
\definecolor{classcolor}{rgb}{0.7, 0.1, 0.2}
\definecolor{functioncolor}{rgb}{0.1, 0.1, 0.8}
\definecolor{keywordcolor}{rgb}{0.6, 0.2, 0.6}

\newcommand{\comment}[1]{\textcolor{commentcolor}{#1}}
\newcommand{\function}[1]{\textcolor{functioncolor}{\textbf{#1}}}

\begin{algorithm}[t]
\caption{\small{Pseudo Code for Next Patch Prediction (NPP)}}
\label{algo:pseduo_code}
\begin{algorithmic}[0]
\small
\STATE from einops import rearrange
 
\STATE \function{class} NPP(nn.Module): 

\STATE \quad \function{def} tensor\_patchify(self, tensor, p): \comment{\#Patch size = p $\times$ p}
% \STATE \quad \quad \comment{\# (B, N, C)-$>$(B ,N/K, K, C) -$>$(B, N/K, C)}
\STATE \quad \quad patches = rearrange(latent, ``b c (h ph) (w pw) -$>$ b (h w) (ph pw) c”,
ph=p, pw=p)

% \STATE \quad \quad patches = F.unfold(tensor, kernel\_size=PS, stride=PS)
\STATE \quad \quad \function{return} patches.mean(dim=1) \comment{\# Group and mean}

\STATE \quad \function{def} label\_patchify(self, label, p): \comment{\#Patch size = p $\times$ p}
% \STATE \quad \quad \comment{\# (B, N)-$>$(B , N/K, K)}
% \STATE \quad \quad label = F.unfold(label, kernel\_size=PS, stride=PS)
\STATE \quad \quad label = rearrange(label, ``b  (h ph) (w pw) -$>$ b (h w) (ph pw) ”,
ph=p, pw=p)
\STATE \quad \quad \function{return} label \comment{\# Group}

\STATE \quad \function{def} forward(self, tokens, labels, global\_step):
\STATE \quad \quad p = self.get\_current\_patch\_size(global\_step): 

\STATE \quad \quad x = self.tok\_emb(tokens) 

\STATE \quad \quad x = self.tensor\_patchify(x, p) 
% \comment{\# (B, N/K, C)}
\STATE \quad \quad \comment {\# Calculate patch positions and RoPE}
\STATE \quad \quad RoPE = self.RoPE\_2d([$h_p$, $w_p$])

\STATE \quad \quad \comment {\# AR model forwarding}
\STATE \quad \quad pred = self.model(x, RoPE) 
% \comment{\# (B,  N/K, Output\_dim)}
\STATE \quad \quad \comment{\# next patch prediction loss}
\STATE \quad \quad  pred = pred.unsqueeze(2).repeat(1, 1, p*p, 1)
\STATE \quad \quad labels = self.label\_patchify(labels, p) 
% \comment{\# (B, N/K, K)}
\STATE \quad \quad loss = nn.CrossEntropy(pred, labels)
\STATE \quad \quad \function{return} loss
\end{algorithmic}
\end{algorithm}

\begin{table*}[t]
\small
\centering
\begin{tabular}{c|lc|c|c|c|c}
\toprule
Type & Model & \#Para. & FID$\downarrow$ & IS$\uparrow$ & Precision$\uparrow$ & Recall$\uparrow$  \\
\midrule
\multirow{2}{*}{GAN}   & BigGAN~\citep{biggan}  & 112M   & 6.95  & 224.5       & 0.89 & 0.38 \\
 & GigaGAN~\citep{gigagan}  & 569M    & 3.45  & 225.5       & 0.84 & 0.61  \\
 % & StyleGan-XL~\citep{stylegan-xl} & 166M    & 2.30  & 265.1       & 0.78 & 0.53   \\
\midrule
\multirow{4}{*}{Diffusion} & ADM~\citep{adm}   & 554M       & 10.94 & 101.0        & 0.69 & 0.63    \\
 & CDM~\citep{cdm}   & $-$       & 4.88  & 158.7       & $-$  & $-$   \\
 & LDM-4~\citep{ldm}  & 400M     & 3.60  & 247.7       & $-$  & $-$  \\
 & DiT-L/2~\citep{dit}  & 458M  & 5.02  & 167.2       & 0.75 & 0.57   \\
\midrule
\multirow{2}{*}{Mask.} & MaskGIT~\citep{maskgit}  & 227M   & 6.18  & 182.1        & 0.80 & 0.51  \\
 & MaskGIT-re~\citep{maskgit} & 227M\    & 4.02  & 355.6        & $-$ & $-$ \\
\midrule

\multirow{2}{*}{VAR} & VAR-$d16$~\citep{VAR} & 310M & 3.30 & 274.40 & 0.84 & 0.51 \\
 & VAR-$d20$~\citep{VAR} & 600M & 2.57 & 302.60 & 0.83 & 0.56 \\
 % & VAR-$d24$-$350epochs$~\citep{VAR} & 1.0B & 2.09 & 312.90 & 0.82 & 0.59  \\
 % & VAR-$d30$-$350epochs$~\citep{VAR} & 2.0B & 1.92 & 323.1 & 0.82 & 0.59 \\
\midrule

\multirow{9}{*}{AR} & VQGAN~\citep{vqgan}  & 227M & 18.65 & 80.4         & 0.78 & 0.26    \\
 & VQGAN~\citep{vqgan}     & 1.4B   & 15.78 & 74.3   & $-$  & $-$     \\
 & VQGAN-re~\citep{vqgan}  & 1.4B  & 5.20  & 280.3  & $-$  & $-$     \\
 & ViT-VQGAN~\citep{vit-vqgan} & 1.7B & 4.17  & 175.1  & $-$  & $-$        \\
 & ViT-VQGAN-re~\citep{vit-vqgan}& 1.7B  & 3.04  & 227.4  & $-$  & $-$     \\
 & RQTran.~\citep{rq}      & 3.8B  & 7.55  & 134.0  & $-$  & $-$     \\
 & RQTran.-re~\citep{rq}     & 3.8B & 3.80  & 323.7  & $-$  & $-$    \\
 & GPT2-re~\citep{vqgan}     & 1.4B & 5.20  & 280.3  & $-$  & $-$  \\
 & Open-MAGVIT2-B~\citep{luo2024open}     & 343M & 3.08  & 258.3  & 0.85  & 0.51  \\
\midrule

\multirow{7}{*}{AR} & LlamaGen-B~\citep{llamagen} & 111M & 5.46 & 193.61 & 0.83 & 0.45\\
 & LlamaGen-L~\citep{llamagen} & 343M & 3.80 & 248.28 & 0.83 & 0.52\\
 & LlamaGen-L-384†~\citep{llamagen}  & 343M & 3.07 & 256.06 & 0.83 & 0.52 \\
 & LlamaGen-XL~\citep{llamagen}  & 775M & 3.39 & 227.08 & 0.81 & 0.54 \\
 & LlamaGen-XL-384†~\citep{llamagen}  & 775M & 2.62 & 244.08 & 0.80 & 0.57 \\
 & LlamaGen-XXL~\citep{llamagen}  & 1.4B & 3.10 & 253.61 & 0.83 & 0.53 \\
 & LlamaGen-XXL-384†~\citep{llamagen}  & 1.4B & 2.34 & 253.90 & 0.80 & 0.59 \\
 % & LlamaGen-3B (cfg=1.65) & 3.1B & 2.18 & 263.33 & 0.81 & 0.58 \\
 % & LlamaGen-3B†~\citep{llamagen}  & 3.1B &  2.18 & 263.33 & 0.81 & 0.58  \\
% \hdashline
\multirow{4}{*}{Ours}
 &  NPP-B  & 111M &4.47 & 229.25  & 0.86 &0.46  \\
 &  NPP-L  & 343M & 2.76 & 266.34 & 0.83& 0.56 \\
&   NPP-XL  & 775M & 2.65 & 281.03  & 0.83 & 0.57 \\
& NPP-XXL  & 1.4B &  2.54& 286.13 & 0.84 & 0.56\\
% &  LlamaGen-3B + NPP ($\lambda$=$2/3$, cfg=2.00) & 3B &  &  &  & \\
\bottomrule
\end{tabular}

\caption{\textbf{Model comparisons on class-conditional ImageNet 256$\times$256 benchmark.} Metrics include Fréchet Inception Distance (FID)~\citep{fid}, Inception Score (IS)~\citep{inception_score}, Precision and Recall.
``$\downarrow$'' or ``$\uparrow$'' indicate lower or higher values are better.
``-re'' means using rejection sampling. ``†'' means the model is trained on $384\times384$ resolution and resized to $256\times256$ for evaluation. 
}
\label{tab:main}
\end{table*}

\section{Experiments}
In this section, we first describe the implementation details of the proposed method in Section~\ref{sec:4.1}. The main results are provided in Section~\ref{sec:4.2}, followed by training cost study and visualization results in Section~\ref{sec:cost} and ~\ref{sec:vis}. We also provide ablation studies on key design choices in Section~\ref{sec:ab}.

\subsection{Implementation Details}
\label{sec:4.1}
\textbf{Benchmark.} We build the Next Patch Prediction based on LlamaGen~\citep{llamagen}  and evaluate it on the class-conditional image generation task using the standard ImageNet1K $256\times256$ generation benchmark~\citep{imagenet}.  

\noindent \textbf{Model Architecture.} For the image encoder, we adopt the same VQGAN tokenizer trained by LlamaGen on ImageNet1K. The tokenizer has a vocabulary size of 16,384 and downsamples the input image at a fixed ratio of $16 \times 16$. For the autoregressive model, we adopt the same setting as LlamaGen. Note that our method does NOT introduce any extra trainable parameters and thereby can be easily extended to other autoregressive models or scaling up to similar tasks.

\noindent \textbf{Training \& Inference Settings.}
All the model are trained for 300 epochs following the same setting of LlamaGen~\citep{llamagen}: base learning rate of $1 \times 10^{-4}$ per 256 batchsize, AdamW optimizer with $\beta_1=0.9,\beta_2 = 0.95$, weight decay $= 0.05$, gradient clipping set to 1.0. To enable smooth transfer between different patch size segments, we set learning rate warmup for the first 1 epoch and linearly decay to $1 \times 10^{-5}$ for the last $1/5$ number of epochs in each segment. The dropout ratio in the autoregressive model backbone is set to 0.1. We also set the class token embedding dropout ratio to 0.1 for classifier-free guidance. For inference, as our method does not modify the inference stage, we follow vanilla next token prediction and adopt the sampling configurations of top-k = 0 (all), top-p = 1.0, and temperature = 1.0, which are the same inference setting as LlamaGen~\citep{llamagen}.

\noindent \textbf{Evaluation Settings.} Following the standard protocols, we sample 50,000 images with trained models to evaluate the Fréchet Inception Distance (FID)~\cite{fid} score, Inception Score (IS)~\cite{inception_score}, Precision and Recall. We follow previous work to use classifier-free guidance during the sampling process. The complete settings of hyper-parameters for each model variant are provided in the appendix.

\begin{figure}[h]
    \centering
    % \vspace{-1em}
    \includegraphics[width=\linewidth]{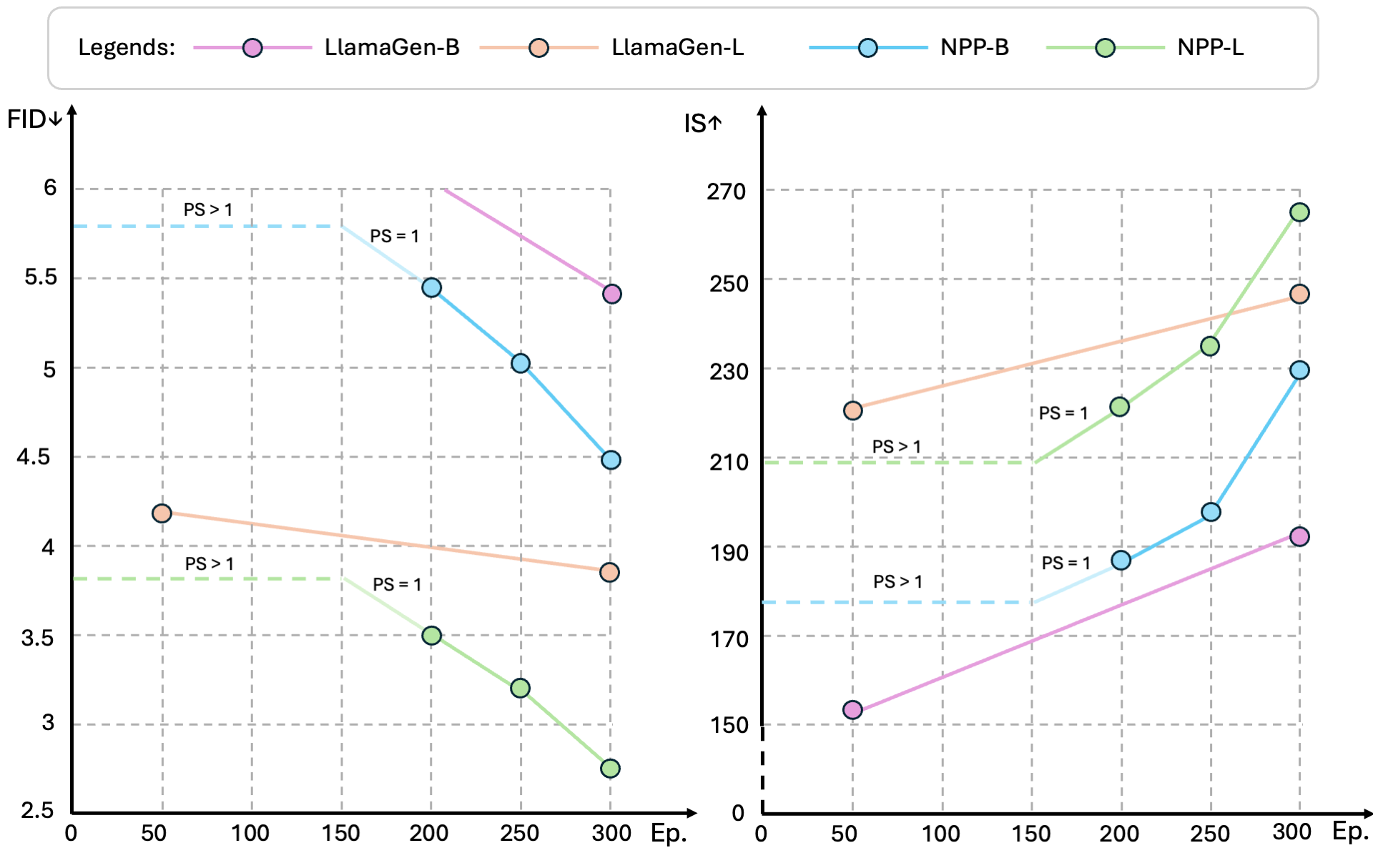}
    \caption{\textbf{Comparison of our method and baseline methods.} The vertical axes are the FID score and IS score. We record the performance curve with the number of epochs as horizontal axes.}
    % \vspace{-1em}
    \label{fig:chart}
\end{figure}
% \vspace{-2em}

\noindent \textbf{Baseline Methods.} 
We choose baseline methods from popular image generation models, including GAN~\citep{biggan,gigagan,stylegan-xl}, Diffusion models~\citep{adm,cdm,ldm,dit}, masked-prediction models~\citep{maskgit} and autoregressive models~\citep{vqgan,vit-vqgan,rq,VAR}. As our method is built upon LlamaGen~\citep{llamagen}, we take it as a strong baseline and mainly compare our method with it.

\begin{table*}[h]
    \small
    \centering
    \begin{tabular}{lc|c|c|cc}
    \toprule
    Model  & \#para. &FID$\downarrow$ &  \quad Cost(GFLOPs)$\downarrow$ \quad & Throughput(imgs/sec)$\uparrow$\\
    \midrule
     LlamaGen-B~\citep{llamagen}   & 111M & 5.46  &   25.06 \textcolor{gray}{$(1.00\times)$}  & $\sim$5888 \textcolor{gray}{$(1.0\times)$}  \\
     NPP-B\quad($\#L=2,\lambda=1/2$) & 111M    &  4.47 & 15.70 \textcolor{greenn}{$(0.63\times)$} & $\sim$7625 \textcolor{greenn}{$(1.3\times)$} \\
    \midrule
     VAR-$d16$~\citep{VAR} & 310M & 3.30 & 105.70 \textcolor{redd}{$(1.27\times)$}  & $\sim$ 1078 \textcolor{redd}{$(0.5\times)$} \\
     LlamaGen-L~\citep{llamagen} & 343M   &3.80 &  83.54 \textcolor{gray}{$(1.00\times)$} & $\sim$ 2201 \textcolor{gray}{$ (1.0\times)$}  \\
     NPP-L\quad($\#L=2, \lambda=1/2$)& 343M     &  2.76 &47.95 \textcolor{greenn}{$(0.57\times)$} &$\sim$ 3469 \textcolor{greenn}{$(1.6\times)$}  \\

     \midrule
     VAR-$d20$~\citep{VAR} & 600M & 2.57 & 204.40  \textcolor{redd}{$(1.06\times)$} & $\sim$ 690 \textcolor{redd}{$(0.7\times)$}\\
     LlamaGen-XL~\citep{llamagen} & 775M  &3.39 & 193.35 \textcolor{gray}{$(1.00\times)$} &$\sim $922 \textcolor{gray}{$ (1.0\times)$}   \\
     LlamaGen-XL-384†~\citep{llamagen} & 775M  &2.62 &  434.11 \textcolor{redd}{$(2.25\times)$} & $\sim$ 410 \textcolor{redd}{$(0.5\times)$}  \\
     NPP-XL\quad($\#L=2, \lambda=1/2$)  & 775M  & 2.65  & 102.78 \textcolor{greenn}{$(0.53\times)$} &$\sim $1613 \textcolor{greenn}{$(1.8\times)$}  \\
    \midrule
     % VAR-$d24$-$350epochs$~\citep{VAR} & 1.0B & 2.09 & 351.29   \\
    LlamaGen-XXL~\citep{llamagen} & 1.4B & 3.10 & 355.72 \textcolor{gray}{$(1.00\times)$} &$\sim $448 \textcolor{gray}{$ (1.0\times)$} \\
    LlamaGen-XXL-384†~\citep{llamagen} & 1.4B & 2.34 & 798.64 \textcolor{redd}{$(2.25\times)$} & $\sim$ 298 \textcolor{redd}{$(0.7\times)$}    \\
     NPP-XXL\quad($\#L=2, \lambda=1/2$) & 1.4B   & 2.54 &  189.11  \textcolor{greenn}{$(0.53\times)$} &$\sim$ 640  \textcolor{greenn}{$ (1.4\times)$}  \\
    % \midrule
    %  VAR-$d30$-$350epochs$~\citep{VAR} & 2.0B & 1.92 & 683.04   \\
    % LlamaGen-3B†~\citep{llamagen} & 3.1B & 2.18& 1754.58   \\
    %  LlamaGen-3B + NPP\quad($\#L=4, \lambda=2/3$) & 3.1B   &  &316.58(319.74) \\

    \bottomrule
    \end{tabular}
    \caption{\textbf{Comparisons training cost on class-conditional ImageNet 256$\times$256 benchmark.}``†'' means the model is trained on $384\times384$ resolution and resized to $256\times256$ for evaluation. We also present the average training throughput per second.}
    \label{tab:cost}
\end{table*}

\subsection{Main results}
\label{sec:4.2}

We compare our method with various baseline works on class-conditional ImageNet 256×256 benchmark and show the results in Table~\ref{tab:main}. Our method achieves state-of-the-art performance on a diverse model size from 100M to 1.4B parameters compared to baseline methods. Specifically, the NPP-L with only 343M parameters achieves a 2.76 FID score, significantly surpassing state-of-the-art AR models with similar parameters including LlamaGen-L-384~\cite{llamagen} (FID 3.07), Open-MAGVIT2-B~\cite{luo2024open} (FID 3.08). It also outperforms the widely used VAR-$d$16~\cite{VAR} (FID 3.30) and the diffusion model DiT-L/2~\cite{dit} (FID 5.02). Moreover, compared with LlamaGen-XL and LlamaGen-XXL, our method consistently outperforms the baseline work trained on 256$\times$256 resolution.

To better compare our method with the strong baseline LlamaGen~\cite{llamagen}, we provide a comprehensive study as shown in Figure~\ref{fig:chart}. We report the detailed evaluation metrics FID and IS as training epochs increase. For the base model and large model, our method consistently outperforms LlamaGen during the training process, improving the FID score and inception score.  In general, the proposed method outperforms the baseline work LlamaGen by improving the image generation quality up to 1.0 FID scores with significantly higher IS scores.

\subsection{Training Cost Study}
\label{sec:cost}

We provide a comprehensive study on the training cost as shown in Table~\ref{tab:cost}. We compare baseline methods including LlamaGen~\cite{llamagen} and VAR~\cite{VAR} with our method across various model sizes. The average computation cost (GFLOPs per batch) and the actual training throughput (images per second) are presented. For models with 100M-300M parameters, NPP reduces the computation cost to $\sim 0.6 \times$ and speeds up the training process by $\sim 1.3 \times$ to $1.6 \times$. Surprisingly, NPP even achieves better generation quality (lower FID scores) with significantly better training efficiency. For large models with 600M-1.4B parameters, our method achieves the best balance between model performance and training efficiency. Specifically, NPP-XL achieves a similar FID score as LlamaGen-XL-384 (2.65 vs 2.62), but with only $\sim 0.25 \times$  training cost and speed up the training process by a $\sim 4 \times$ model throughput.

\begin{figure}[h]
    \centering
    \includegraphics[width=\linewidth]{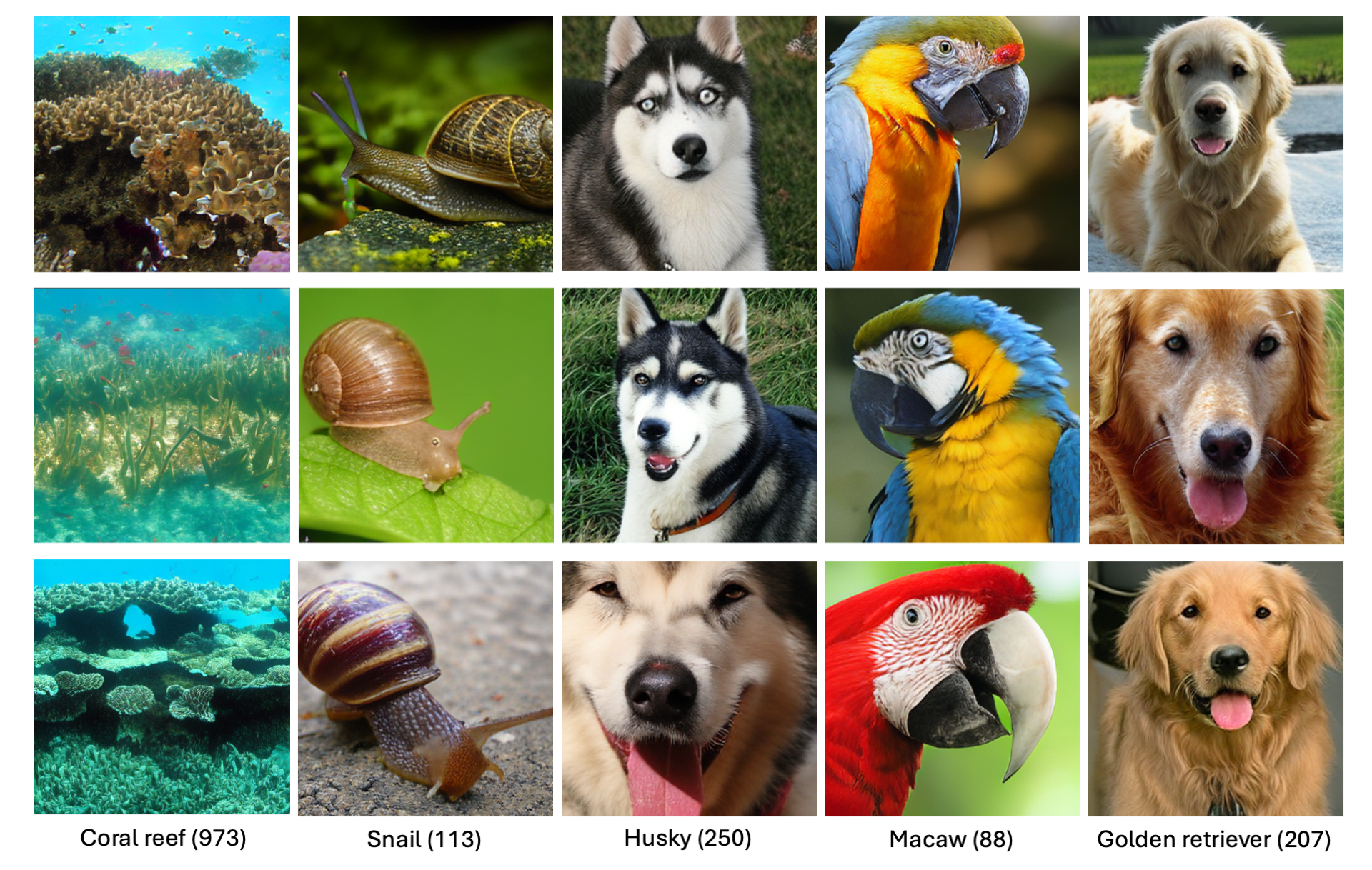}
    % \vspace{-1.5em}
    \caption{\textbf{Generation results.} Please zoom in to view.}
    \label{fig:gen}
\end{figure}
% \vspace{-1.5em}

\subsection{Generation Results}
\label{sec:vis}

 In Figure~\ref{fig:gen}, we present generation results by NPP  on ImageNet 256$\times$256 benchmark. Our NPP is capable of generating high-quality images with diversity and fidelity.
More generation results are provided in the appendix.

\begin{table*}[ht]
    \centering
    \small
    \begin{subtable}[t]{\linewidth}
    \centering
    \begin{tabular}{lc|cccc|c}
    \toprule
    Model &\#para.  &FID$\downarrow$ & IS$\uparrow$ & Precision$\uparrow$ & Recall$\uparrow$ & Training Cost$\downarrow$  \\
    \midrule
     LlamaGen-B~\citep{llamagen} \textcolor{gray}{($PS=1\times1$)} & 111M  & 5.46 & 193.61 & 0.83 & 0.45 & $1.0 \times$  \\
     NPP-B \quad ($PS=2\times2$) & 111M   &4.47 & 229.25  & 0.86 & 0.46 & $0.625 \times$ \\
     NPP-B \quad ($PS=4\times4$) & 111M   & 4.92& 222.81  & 0.86 &0.45  & $0.531 \times$ \\
    \midrule
     LlamaGen-L~\citep{llamagen} \textcolor{gray}{($PS=1\times1$)}& 343M    &3.80 & 248.28 & 0.83 & 0.52 & $1.0 \times$  \\
     NPP-L \quad ($PS=2\times2$) & 343M   &2.76 & 266.34  & 0.83 & 0.56  & $0.625 \times$ \\
     NPP-L \quad ($PS=4\times4$) & 343M    & 2.89  & 262.80 & 0.83 & 0.55 & $0.531 \times$ \\

    \bottomrule
    \end{tabular}
    \caption{Comparisons of models trained with different patch sizes.}
    \label{tab:ps}
    \end{subtable}

    \begin{subtable}[h]{\linewidth}
    \centering
    \begin{tabular}{lc|cccc|c}
    \toprule
    Model &\#para. &FID$\downarrow$ & IS$\uparrow$ & Precision$\uparrow$ & Recall$\uparrow$ & Training Cost$\downarrow$  \\
    \midrule
    %  LlamaGen-B~\citep{llamagen}   & 5.46 & 193.61 & 0.83 & 0.45 & $1.0 \times$  \\
    %  LlamaGen-B + NPP\quad($\lambda=1/2$)   &5.17 & 196.39  & 0.84 & 0.46 & $0.625 \times$ \\
    % \midrule
     LlamaGen-L~\citep{llamagen} \textcolor{gray}{($\lambda=0$)}& 343M    &3.80 & 248.28 & 0.83 & 0.52 & $1.0 \times$  \\
     NPP-L\quad($\lambda=1/2$) & 343M   &2.76 & 266.34  & 0.83 & 0.56  & $0.625 \times$ \\
     NPP-L\quad($\lambda=2/3$) & 343M   & 2.79 & 263.75  & 0.83 & 0.55&$0.5 \times$  \\
     NPP-L\quad($\lambda=3/4$) & 343M  &2.81 & 262.22& 0.83 & 0.55  &$0.43 \times$  \\
     NPP-L\quad($\lambda=4/5$) & 343M   & 2.92& 260.68 &0.83 &0.55  &$0.4 \times$  \\
        
    \bottomrule
    \end{tabular}
    \caption{Comparisons of models trained with different segment factor $\lambda$.} 
    \label{tab:factor}
    \end{subtable}

    \begin{subtable}[h]{\linewidth}
        \centering
    \begin{tabular}{lc|cccc|c}
    \toprule
    Model&\#para.  &FID$\downarrow$ & IS$\uparrow$ & Precision$\uparrow$ & Recall$\uparrow$ & Training Cost$\downarrow$  \\
    \midrule
     LlamaGen-B~\citep{llamagen} \textcolor{gray}{($\#L=1$)} & 111M   & 5.46 & 193.61 & 0.83 & 0.45 & $1.0 \times$  \\
     NPP-B\quad($\#L=2$) & 111M   &4.47 & 229.25  & 0.86 & 0.46 & $0.625 \times$ \\
     NPP-B\quad($\#L=3$)  & 111M  &4.62 & 231.57  & 0.86 & 0.46 & $ 0.578 \times$ \\
     NPP-B\quad($\#L=4$) & 111M   &4.68 & 228.31  & 0.86 &0.46  & $0.572\times$ \\
    \midrule
     LlamaGen-L~\citep{llamagen} \textcolor{gray}{($\#L=1$)} & 343M    &3.80 & 248.28 & 0.83 & 0.52 & $1.0 \times$  \\
     NPP-L\quad($\#L=2$) & 343M   &2.76 & 266.34  & 0.83 & 0.56  & $0.625\times$ \\
     NPP-L\quad($\#L=3$) & 343M   &2.79 & 264.30  & 0.83 & 0.56 & $ 0.578 \times$ \\
     NPP-L\quad($\#L=4$) & 343M   &2.84 & 258.60  & 0.83 & 0.56  & $0.572\times$ \\

    \bottomrule
    \end{tabular}
    \caption{Comparisons of models trained with different numbers of patch level.} 
    \label{tab:level}
    \end{subtable}

    \caption{\textbf{Ablation studies on key design choices.} We evaluate the models on class-conditional ImageNet 256$\times$256 benchmark and report the FID score, IS score, Precision, and Recall, along with the theoretical training cost.}
    \label{tab:abl}
    
\end{table*}

\subsection{Ablation Studies}
\label{sec:ab}
 
\noindent \textbf{Effect of Patch Size.}
We study the effect of different patch sizes and present the results in Table~\ref{tab:ps}. In this experiment, we modify the multi-scale grouping strategy to skip intermediate patch size and set the segment scheduling factor $\lambda=1/2$. The models are ablated with different patch sizes adopted in the first $1/2$ number of training epochs. We observe NPP with different patch sizes consistently outperforms LlamaGen. However, with a larger patch size such as $PS=4\times4$, the learned knowledge cannot be smoothly transferred to the case with $PS=1\times1$, leading to a slight performance drop where FID scores were reduced by 0.45 for NPP-B and 0.13 for NPP-L.  Therefore, we choose patch size $PS=2\times2$ as the default setting.

\noindent \textbf{Effect of Segment Scheduling Factor $\lambda$.}
We provide a study on the effect of different segment scheduling factors adopted in the proposed multi-scale patch grouping strategy as shown in Table~\ref{tab:factor}. In this study, the multi-scale patch grouping strategy is disabled and the patch size is set to $PS=2\times2$. $\lambda$ factors are scanned from 1/2 to 4/5. We observe that a larger $\lambda$ factor results in lower training computational cost but with slight performance degradation that FID scores are increased from 2.76 to 2.92.
Hence, to balance training efficiency and model performance, we set $\lambda = 1/2$ by default.

\noindent \textbf{Effect of Multi-scale Patch Grouping Strategy.}
We present a study on the effect of the multi-scale patch grouping strategy as shown in Table~\ref{tab:level}. In this experiment, we set $\lambda=1/2$ and compare different numbers of patch levels $\#L$. Experiments show this strategy makes a trade-off between training computational cost and image generation quality. Moreover, with this strategy, the training process ends with vanilla NTP where the patch size is 1×1, thus preserving the original inference stage without modifications.

% \begin{figure*}[ht]
%     \centering
%     \includegraphics[width=1.0\linewidth]{vis.png}
%     \vspace{-2em}
%     \caption{Generation results on ImageNet 256$\times$256 benchmark.}
%     \label{fig:gen}
% \end{figure*}

% \input{ICCV2025-Author-Kit-Feb/sec/4.5_Limitations}
\section{Conclusion}

In this work, we introduce a novel Next Patch Prediction paradigm that improves autoregressive image generation quality and efficiency by grouping and aggregating image tokens into high-density patch tokens. We further introduce a multi-scale patch strategy to seamlessly bridge the Next Patch Prediction with the vanilla next token prediction paradigm. Our approach reduces the computational cost to $\sim 0.6\times$ while improving image generation quality by up to 1.0 FID score on the ImageNet benchmark. We highlight that our method retains the original autoregressive model architecture without introducing additional trainable parameters or custom image tokenizers, thereby making the next patch prediction paradigm seamlessly adapted to various autoregressive models addressing image generation tasks.

% \clearpage

{
    \small
    \bibliographystyle{ieeenat_fullname}
    \bibliography{main}
}

\end{document}

\title{Appendix for `` Next Patch Prediction for AutoRegressive Visual Generation "}  % **** Enter the paper title here

\maketitle
\thispagestyle{empty}
\appendix

The appendix includes the following additional information:
\begin{itemize}
    \item Section~\ref{sec:A} provides more implementation details for NPP.
    \item Section~\ref{sec:B} presents limitations and future works.
    \item Section~\ref{sec:C} provides more generation results.

\end{itemize}

\section{Implementation Details}
\label{sec:A}

 For image generation experiments, we train our method on the ImageNet-1K training set, which consists of 1,281,167 images across 1,000 object classes. Following the setting in LlamaGen, we pre-tokenize the entire training set using their VQGAN tokenizer and process the data through a ten-crop transformation for augmentation. For inference, we adopt classifier-free guidance to improve generation quality following LlamaGen. The detailed training and sampling hyper-parameters are listed in Tab.~\ref{tab:img_params}.

\begin{table}[h]
\centering
\begin{tabular}{p{0.45\columnwidth}|p{0.45\columnwidth}}
% \toprule
config & value \\
\Xhline{1.2pt}
\multicolumn{2}{c}{\textit{training hyper-params}} \\
\Xhline{0.8pt}
optimizer & AdamW \\
learning rate & 1e-4  \\
weight decay & 5e-2 \\
optimizer momentum & (0.9, 0.95) \\
batch size & 256 \\
learning rate schedule &constant with linear decay  \\
ending learning rate & 1e-5 \\
total epochs & 300 \\
warmup epochs & 5 \\
precision & bfloat16 \\
max grad norm & 1.0 \\
dropout rate & 0.1 \\
attn dropout rate & 0.1 \\
class label dropout rate & 0.1 \\
\Xhline{0.8pt}
\multicolumn{2}{c}{\textit{sampling hyper-params}} \\
\Xhline{0.8pt}
temperature & 1.0\\
 top-k &  0 (all) \\
 top-p & 1.0 \\
guidance scale & 2.0 (B) / 1.75 (L, XL) / 1.635 (XXL) \\
\Xhline{0.8pt}
\end{tabular}
\caption{Detailed Hyper-parameters for Image Generation.}
\label{tab:img_params}
\end{table}

\section{Limitations and Future Works}
\label{sec:B}
Currently, due to limited computational resources, we are unable to provide experimental results on text-to-image tasks or develop a unified vision-language model. However, our method retains the original autoregressive model architecture without introducing additional trainable parameters or specifically designing a custom image tokenizer. We believe it can be seamlessly adapted to various autoregressive models addressing similar tasks, offering a baseline framework for future exploration. 

For future works, exploring the NPP paradigm for autoregressive video generation is an interesting direction, where the input sequence length is significantly longer, resulting in higher computational costs. Another potential research direction is to explore more effective methods for obtaining patch embeddings. In this work, we adopt a simple intra-patch averaging operation, which is straightforward but may be limited in its capacity to capture fine-grained details. Designing more powerful and expressive patch extraction methods could further improve training efficiency and enhance generation quality.

% . By leveraging our method, we anticipate improved efficiency and scalability in handling the extended sequences inherent in video data. 
% Another potential research direction is to explore more effective methods for obtaining patch embeddings. In this work, we adopt a simple intra-patch averaging operation, which is straightforward but may be limited in its capacity to capture fine-grained details. Designing more powerful and expressive patch extraction methods could further improve training efficiency and enhance generation quality.

\section{Additional Generation Results}
\label{sec:C}
\begin{figure}[h]
    \centering
    \includegraphics[width=0.75\linewidth]{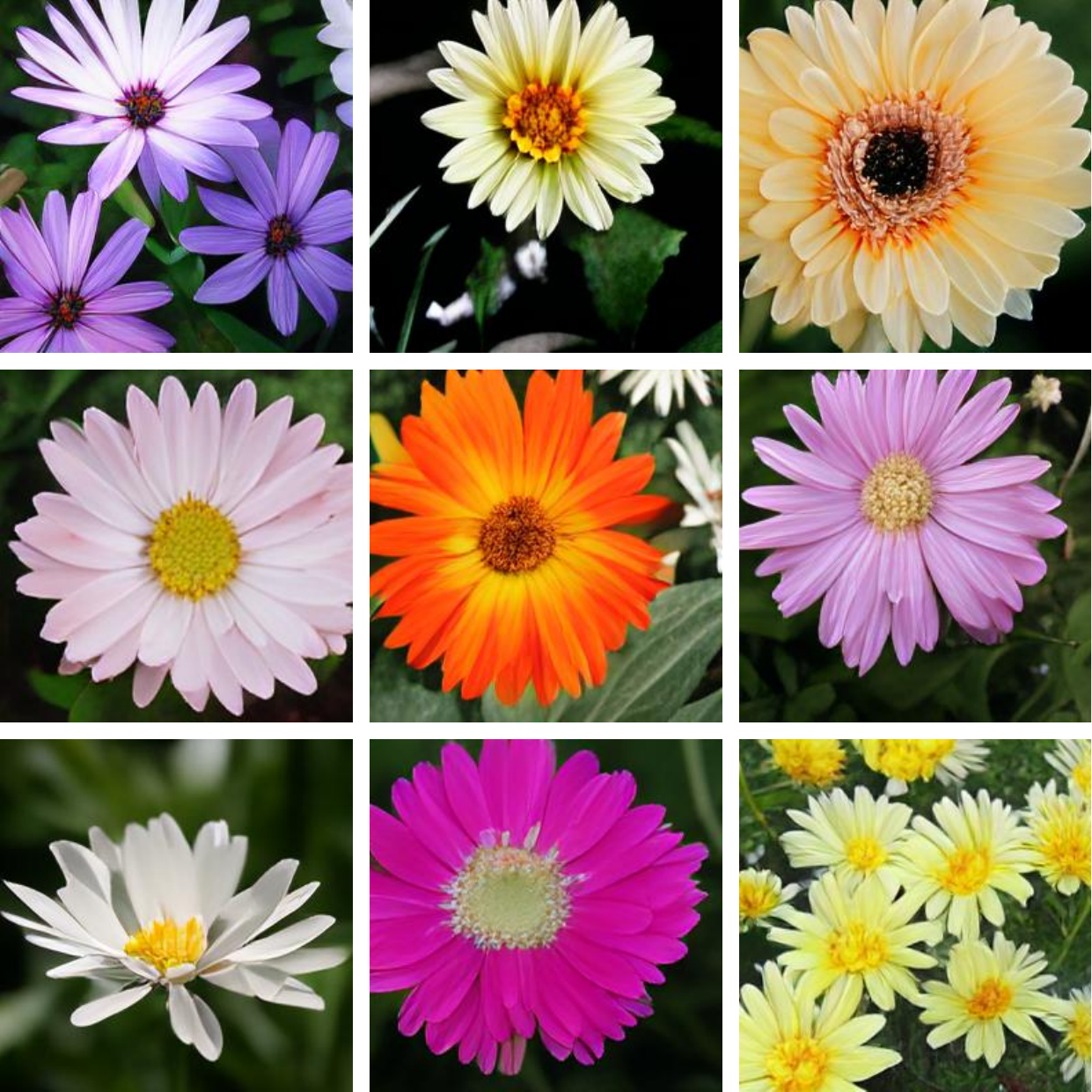}
    \caption{Generation results of daisy (985) images.}
    % \label{fig:enter-label}
\end{figure}
\vspace{-1em}
\begin{figure}[h]
    \centering
    \includegraphics[width=0.75\linewidth]{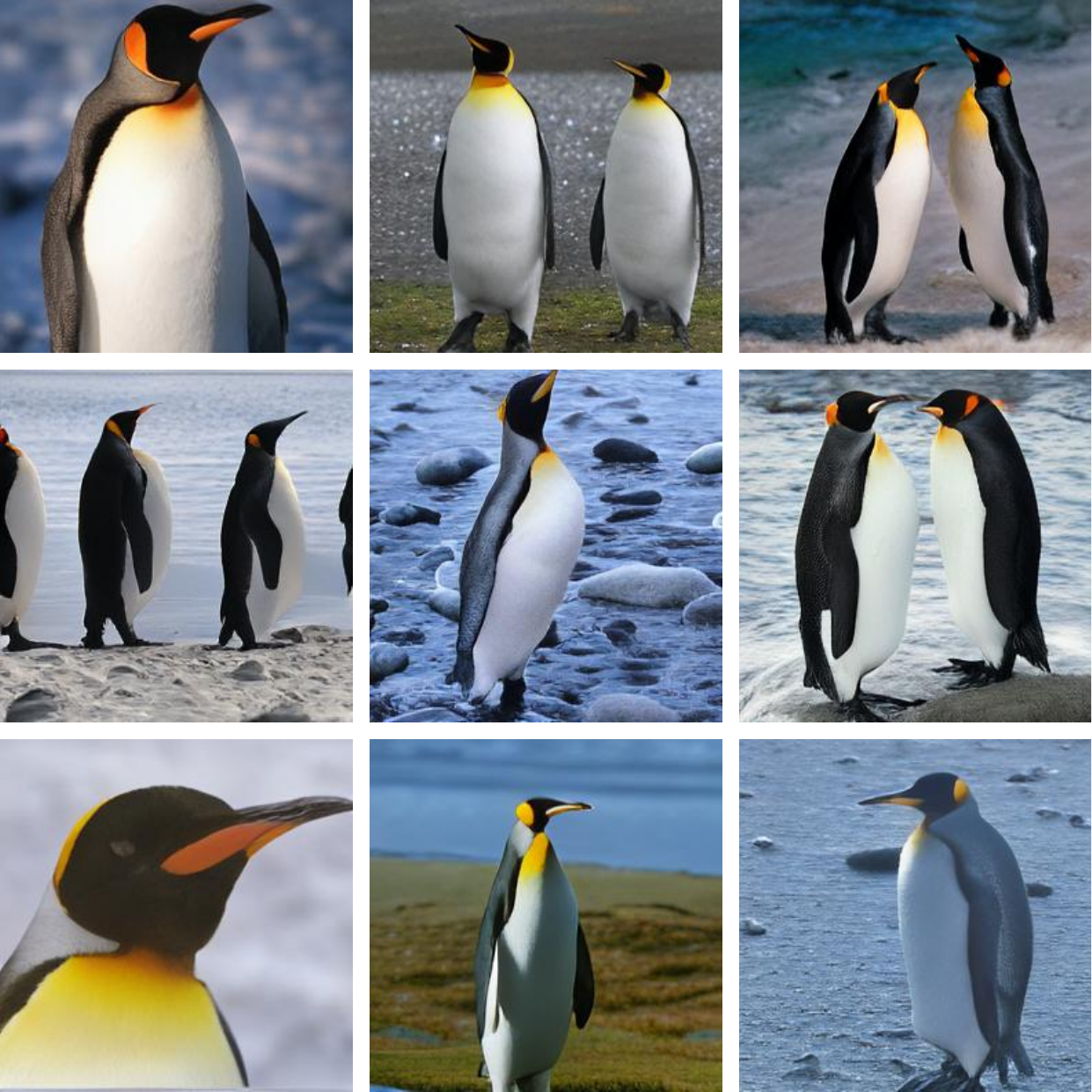}
    \caption{Generation results of king penguin (145) images.}
    % \label{fig:enter-label}
\end{figure}
\begin{figure}[h]
    \centering
    \includegraphics[width=0.75\linewidth]{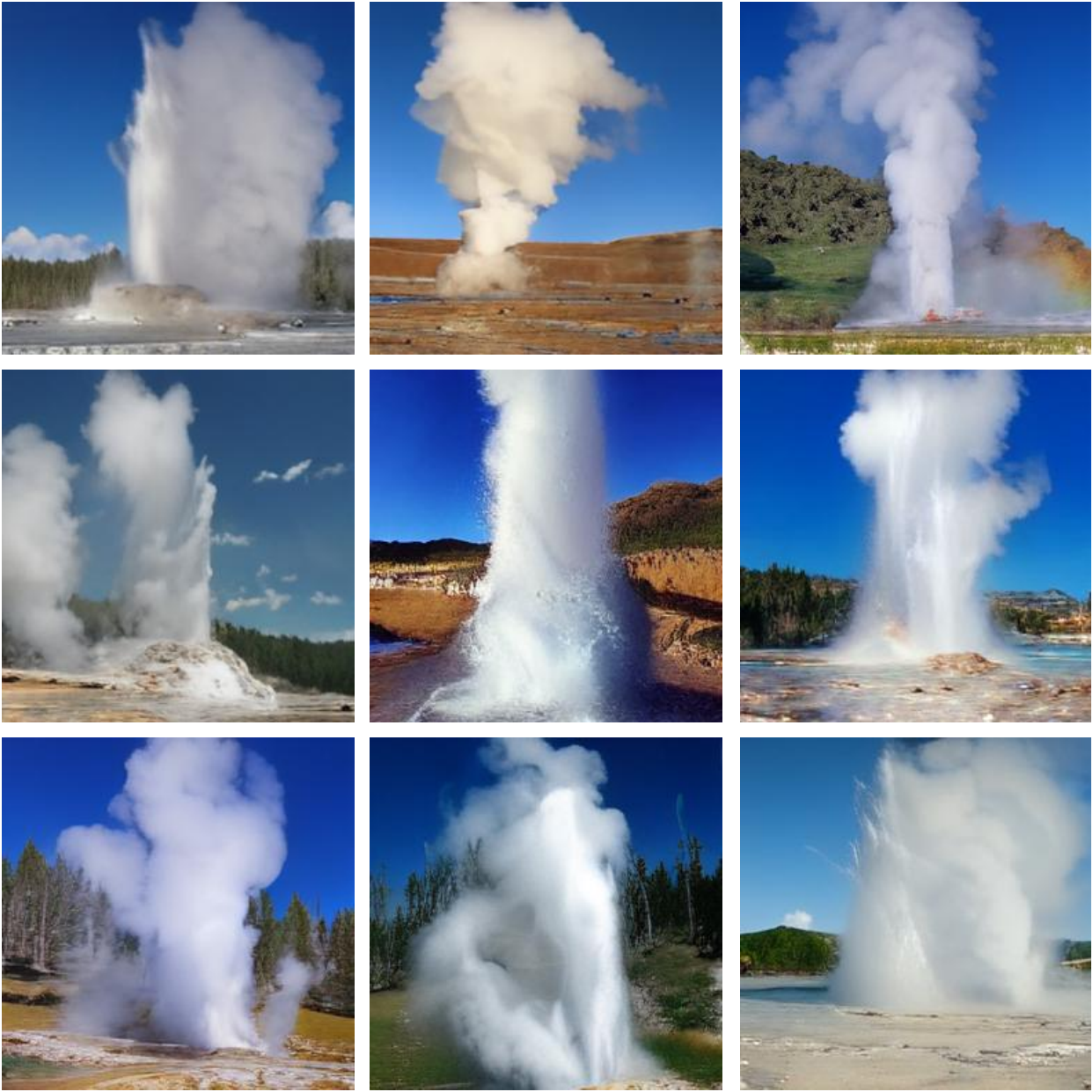}
    \caption{Generation results of geyser (974) images.}
    % \label{fig:enter-label}
\end{figure}

\begin{figure}[h]
    \centering
    \includegraphics[width=0.75\linewidth]{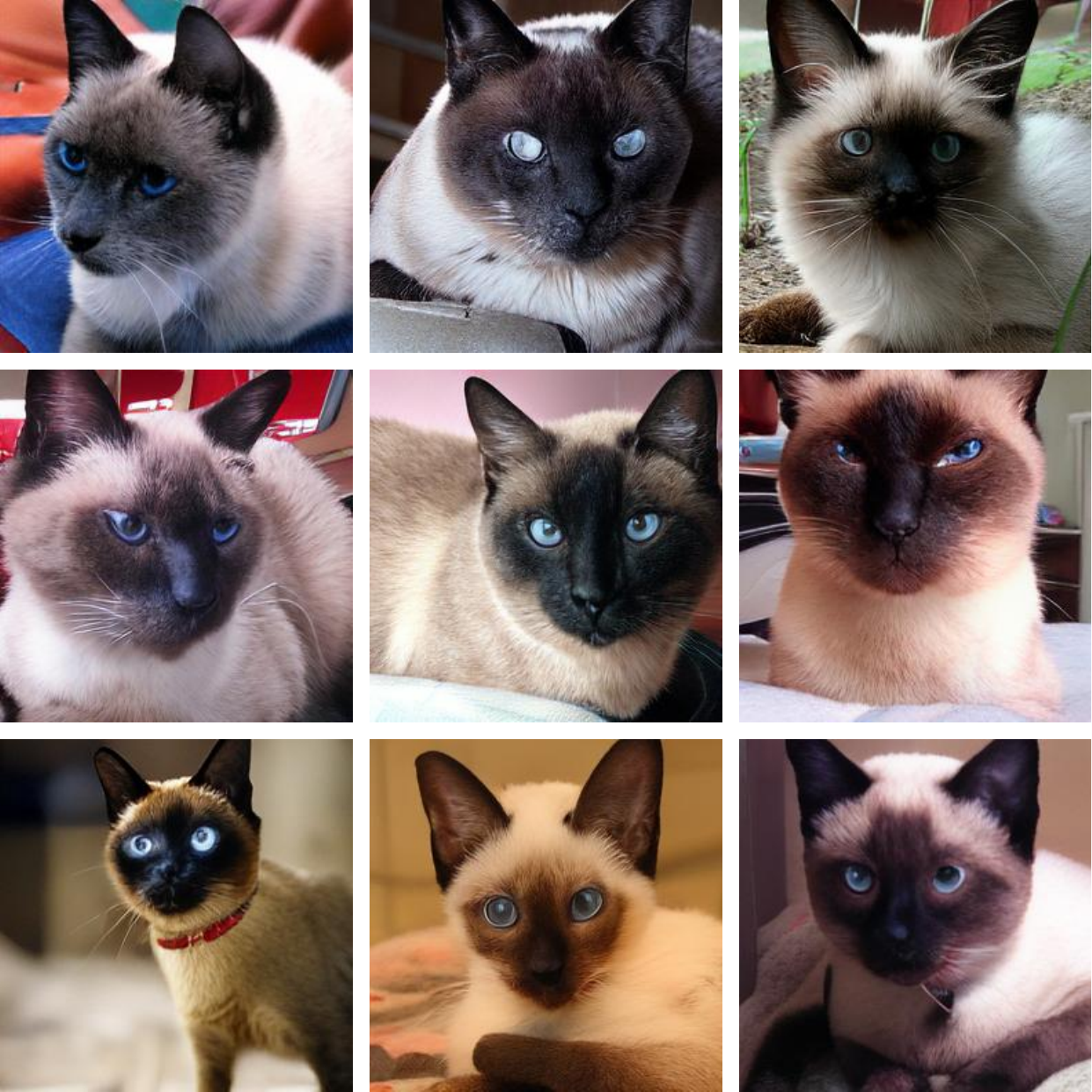}
    \caption{Generation results of Siamese cat (284) images.}
    % \label{fig:enter-label}
\end{figure}
\begin{figure}[h]
    \centering
    \includegraphics[width=0.75\linewidth]{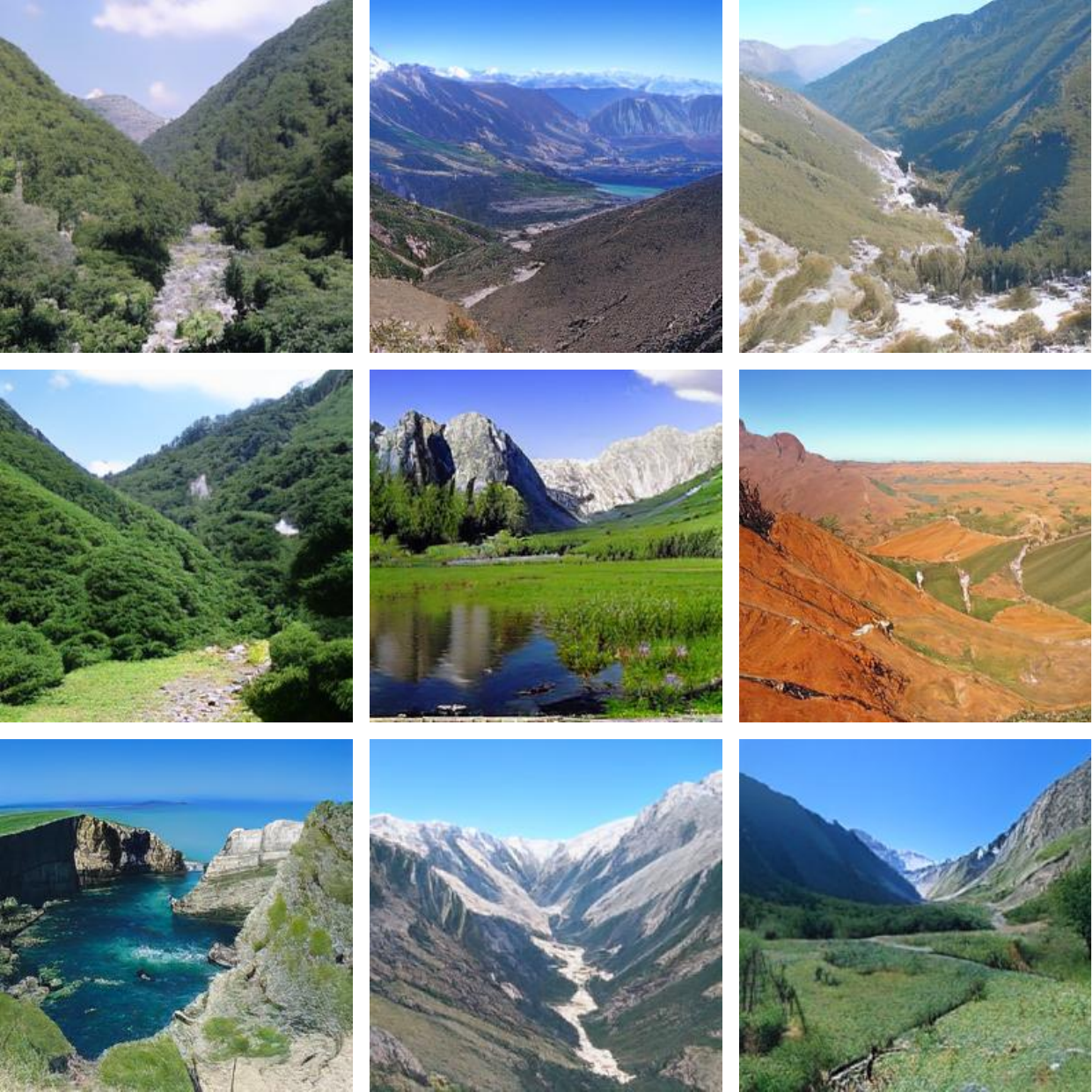}
    \caption{Generation results of valley (979) images.}
    % \label{fig:enter-label}
\end{figure}
\begin{figure}[h]
    \centering
    \includegraphics[width=0.75\linewidth]{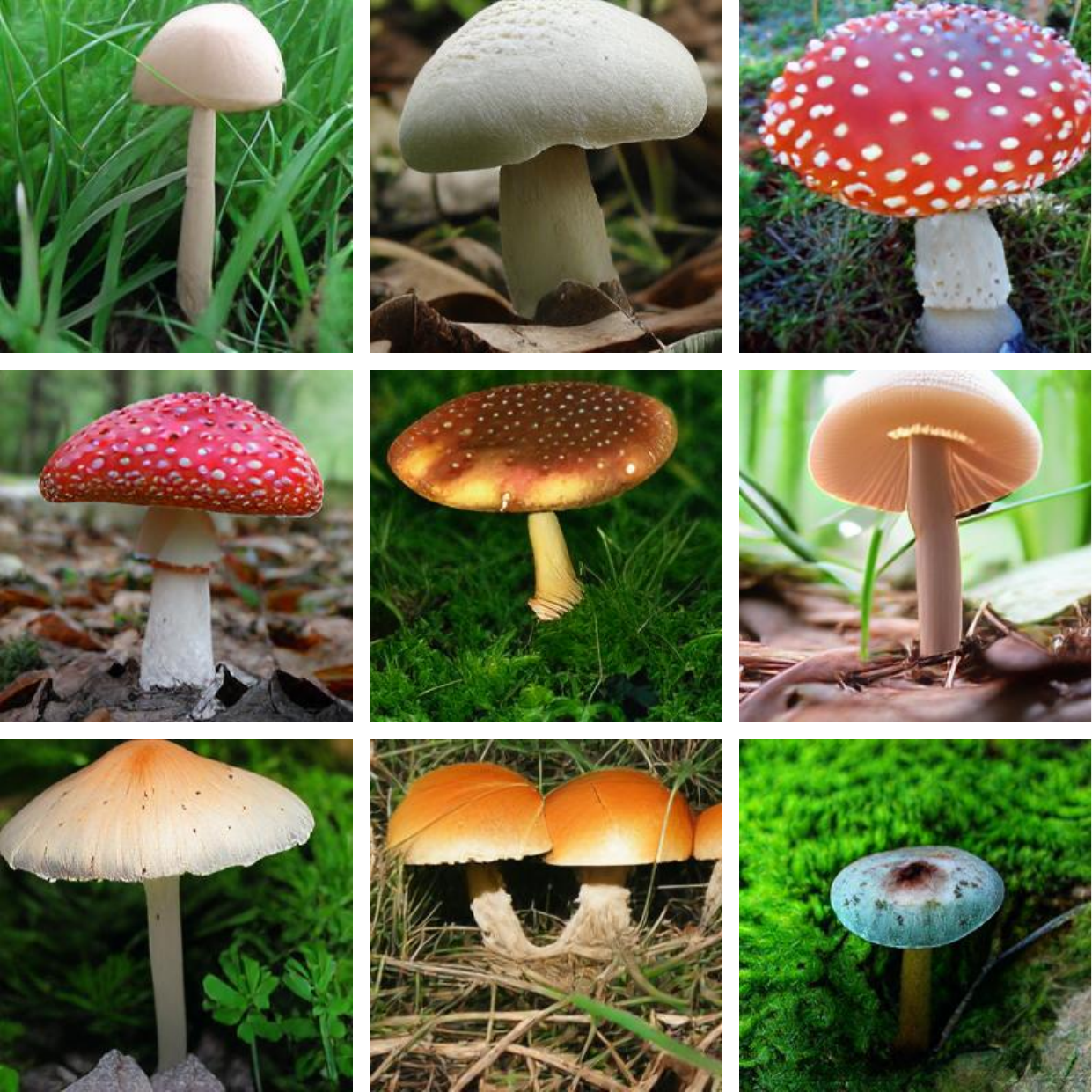}
    \caption{Generation results of mushroom (947) images.}
    % \label{fig:enter-label}
\end{figure}
\begin{figure}[h]
    \centering
    \includegraphics[width=0.75\linewidth]{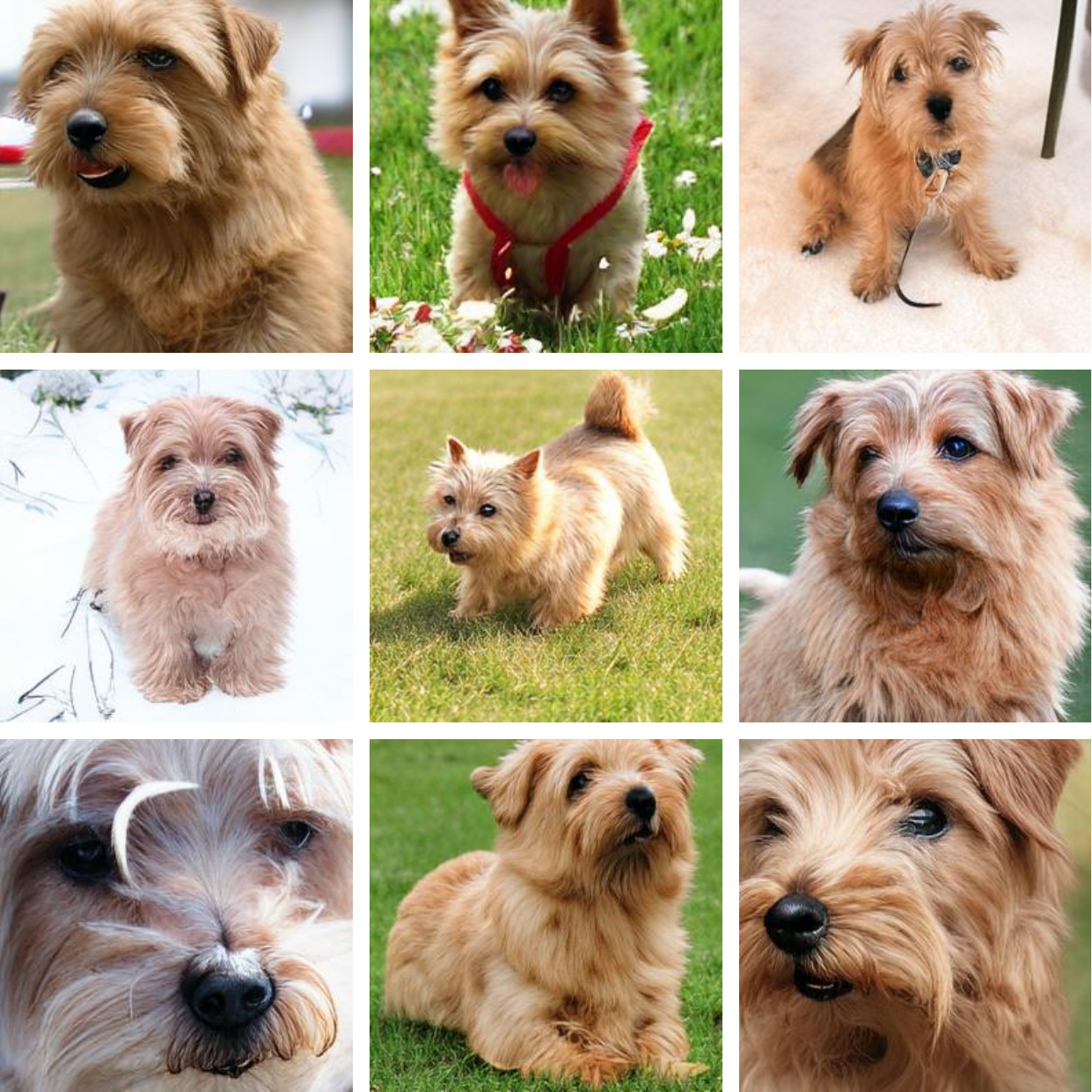}
    \caption{Generation results of Norfolk terrier (185) images.}
    % \label{fig:enter-label}
\end{figure}
\begin{figure}[h]
    \centering
    \includegraphics[width=0.75\linewidth]{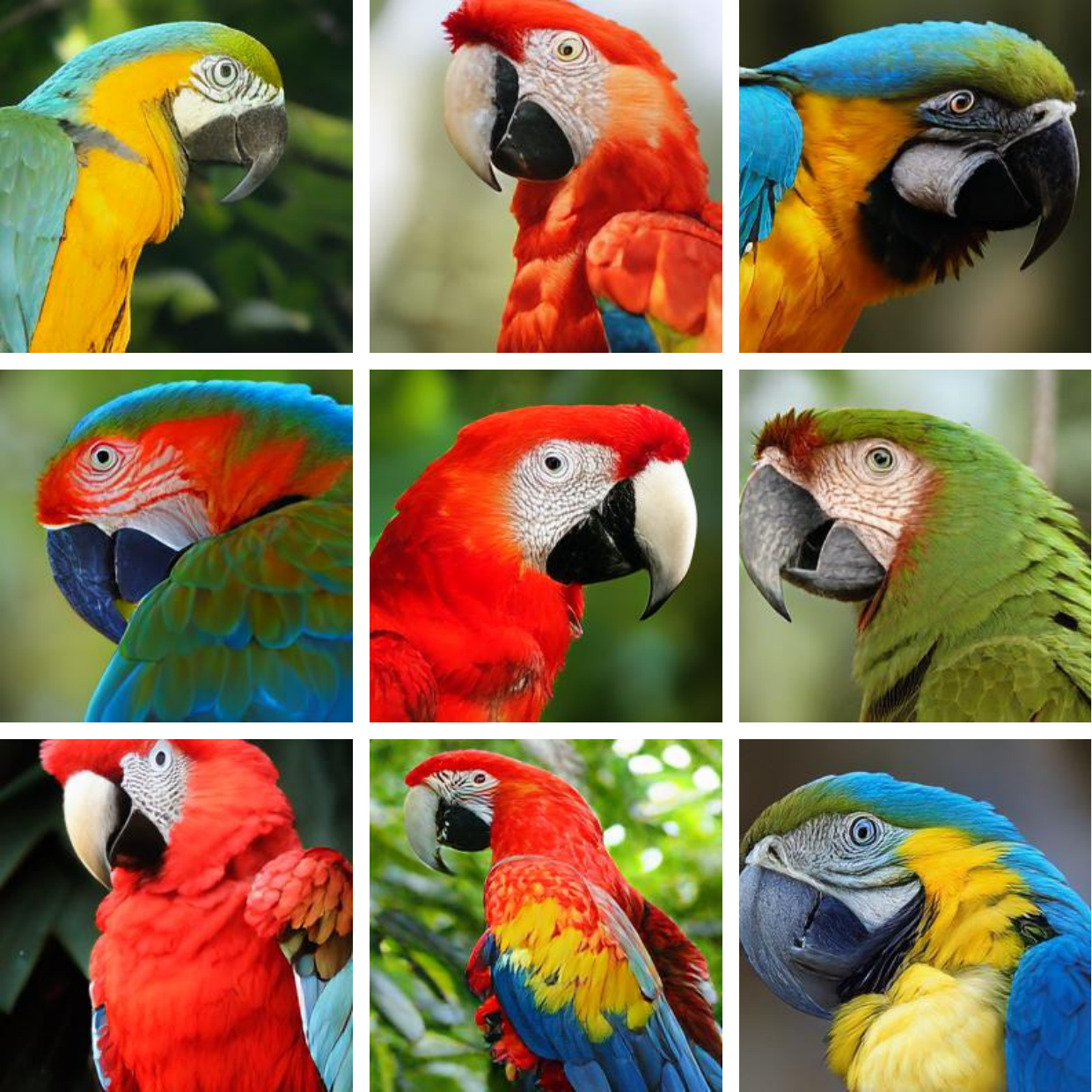}
    \caption{Generation results of macaw (88) images.}
    % \label{fig:enter-label}
\end{figure}
%%%%%%%%% REFERENCES
% {
%     \small
%     \bibliographystyle{ieeenat_fullname}
%     \bibliography{main}
% }